\documentclass[10pt,twocolumn,letterpaper]{article}

\usepackage{iccv}
\usepackage{times}
\usepackage{epsfig}
\usepackage{graphicx}
\usepackage{caption}
\usepackage{amsmath}
\usepackage{amssymb}
\usepackage{booktabs}  
\usepackage{float}
\usepackage{dsfont}
\usepackage[section]{placeins}
\usepackage{overpic}
\usepackage{wrapfig}
\usepackage[dvipsnames]{xcolor}
\usepackage[accsupp]{axessibility}
\usepackage{color}
\definecolor{green}{rgb}{0, 0.5, 0}
\definecolor{orange}{rgb}{0.8, 0.6, 0.2}
\definecolor{red}{rgb}{1.0, 0.0, 0.0}
\definecolor{teal}{rgb}{0.0, 0.4, 0.4}
\definecolor{purple}{rgb}{0.65,0,0.65}
\definecolor{saffron}{rgb}{0.95,0.75,0.2}
\definecolor{turquoise}{rgb}{0.0,0.5,0.5}
\definecolor{black}{rgb}{0.0, 0.0, 0.0}
\definecolor{gray}{rgb}{0.5, 0.5, 0.5}

\newcommand{\ours}{SKED}

\newcommand{\overbar}[1]{\mkern 1.5mu\overline{\mkern-1.5mu#1\mkern-1.5mu}\mkern 1.5mu}
\newcommand{\inputcoord}{\mathbf{p}}
\newcommand{\inputray}{\hat{\mathbf{r}}}
\newcommand{\outbasedensity}{\sigma_o}
\newcommand{\outbasecolor}{\mathbf{c}_o}
\newcommand{\outeditdensity}{\sigma_e}
\newcommand{\outbasealpha}{\alpha_o}
\newcommand{\outeditalpha}{\alpha_e}
\newcommand{\outeditcolor}{\mathbf{c}_e}
\newcommand{\coord}{\mathbf{p}_i}
\newcommand{\pixel}{\mathbf{x}_i}
\newcommand{\preservationweight}{w_i}
\newcommand{\reals}{\mathds{R}}

\newcommand{\basenerf}{F_o}
\newcommand{\basenerfFull}{F_o: (\inputcoord, \inputray; \theta) \to (\outbasecolor, \outbasedensity)}
\newcommand{\edittednerf}{F_e}
\newcommand{\edittednerfFull}{F_e(\inputcoord, \inputray; \phi)}
\newcommand{\sensitivity}{\beta}

\newcommand{\rv}[1]{}

\newcommand{\ts}{@{\hskip 1\tabcolsep}}

\newcommand{\tsm}{@{\hskip 2\tabcolsep}}

\newcommand{\Lstroke}{%
  \text{\ooalign{\hidewidth\raisebox{0.2ex}{--}\hidewidth\cr$\mathcal{L}$\cr}}%
}

\usepackage[pagebackref=true,breaklinks=true,letterpaper=true,colorlinks,bookmarks=false]{hyperref}

 \iccvfinalcopy 

\def\iccvPaperID{1206} 

\ificcvfinal\fi

\begin{document}

\title{SKED: Sketch-guided Text-based 3D Editing}

\author{Aryan Mikaeili$^{1}$ \hspace{6mm}
Or Perel$^{2}$ \hspace{6mm}
Mehdi Safaee$^{1}$ \hspace{6mm}
Daniel Cohen-Or$^{3}$\hspace{6mm}
Ali Mahdavi-Amiri$^{1}$\\\\
$^1$Simon Fraser University \hspace{0.25cm}
$^2$NVIDIA \hspace{0.25cm}
$^3$Tel Aviv University
\vspace{4mm}
}

\twocolumn[{%
\maketitle
\vspace{-14mm}
\begin{figure}[H]
\hsize=\textwidth
\centering
\includegraphics[width=\textwidth]{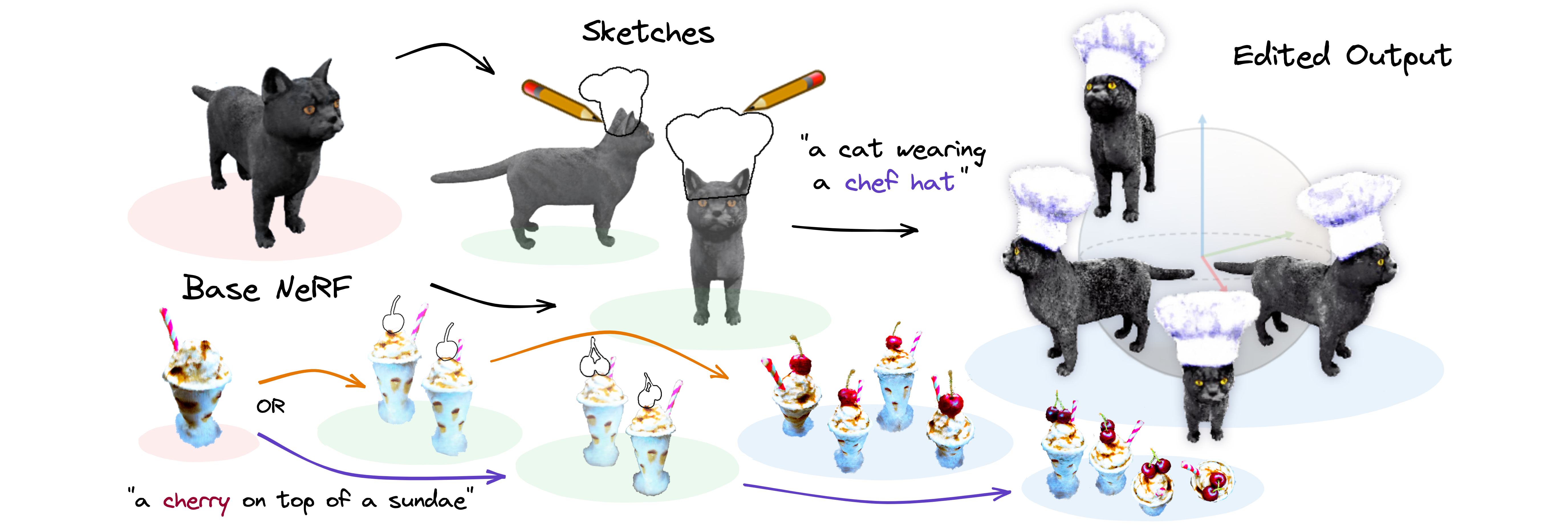}
\captionsetup{font=small}
\caption{Examples of our \textbf{Sketch-guided}, \textbf{Text-based} 3D editing method. Taking a pretrained Neural Radiance Field as input, multiview sketches determining the coarse region of edit and a text-prompt, our method is able to generate a localized, meaningful edit.
}
\vspace{0.5cm}
\label{fig:teaser}
\end{figure}
}]

\ificcvfinal\thispagestyle{empty}\fi

\begin{abstract} 
    Text-to-image diffusion models are gradually introduced into computer graphics, recently enabling the development of Text-to-3D pipelines in an open domain. However, for interactive editing purposes, local manipulations of content through a simplistic textual interface can be arduous. Incorporating user guided sketches with Text-to-image pipelines offers users more intuitive control. 
    Still, as state-of-the-art Text-to-3D pipelines rely on optimizing Neural Radiance Fields (NeRF) through gradients from arbitrary rendering views, conditioning on sketches is not straightforward. In this paper, we present SKED, a technique for editing 3D shapes represented by NeRFs. Our technique utilizes as few as two guiding sketches from different views to alter an existing neural field. The edited region respects the prompt semantics through a pre-trained diffusion model. To ensure the generated output adheres to the provided sketches, we propose novel loss functions to generate the desired edits while preserving the density and radiance of the base instance. We demonstrate the effectiveness of our proposed method through several qualitative and quantitative experiments. \href{https://sked-paper.github.io/}{https://sked-paper.github.io/}
    
\end{abstract}

\section{Introduction}

\label{sec:intro}


Art is a reflection of the figments of human imagination. 
While many are limited in their practical creative capabilities, by pushing the boundaries of digital media, new ways can be created for casual artists and experts alike to express their ideas. At the same time, current neural generative art takes away much of the control from humans. In this work, we attempt to take a step towards restoring some of that control, enabling neural networks to complement users and naturally extend their skills rather than taking hold over the generative process.



The field of image synthesis has been significantly propelled by neural generative models, particularly by the latest text-to-image models that predominantly rely on large language-image models ~\cite{balaji2022eDiff-I, ramesh2022dalle, rombach2021highresolution, imagen2022saharia}. These models have revolutionized the field of computer vision, as they can produce astonishing visual outcomes from text prompts alone.

The ability of text-to-image models has sparked a wave of editing methods that utilize these models. Many of these techniques rely on prompt editing ~\cite{ fu2022shapecrafter, hertz2022prompt, kawar2022imagic,lin2022magic3d,mokady2022null, poole2022dreamfusion}. Nevertheless, simplifying the interface to text alone means users lack the necessary level of granularity to produce their exact desired outcomes.
Sketch-guided editing, on the other hand, provides intuitive control that aligns with user's conventional drawing and painting skills. By incorporating user-guided sketches into text-to-image models, powerful editing systems can be created, offering a high degree of flexibility and fine-grained control for manipulating visual content~\cite{zhang2023controlnet, voynov2022sketch}.

Although sketch-guided and text-driven methods have proven successful in generating and manipulating 2D images \cite{meng2022sdedit, voynov2022sketch, cheng2023wacv}, it immediately raises the intriguing question of whether a similar approach could be developed to edit 3D shapes. 
Since direct text-to-3D models require an abundance of data to scale, state-of-the-art 3D generative models, such as DreamFusion~\cite{poole2022dreamfusion} and Magic3D~\cite{lin2022magic3d}, which build on the capabilities of text-to-image models, may be considered as an alternative.
However, maintaining control via conditioning with such models remains a challenging task, as these generative pipelines optimize a Neural Radiance Field (NeRF) \cite{mildenhall2020nerf} by amortizing gradients from a multitude of 2D views. In particular, providing consistent sketches across all possible views presents a hurdle for users. Instead, a plausible user interface should act with guidance from as few views as possible, e.g. up to two or three.

In this paper, we present \textbf{SKED}, a \textbf{SK}etch-guided 3D \textbf{ED}iting technique. Our method acts on reconstructed or generated NeRF models. We assume a text prompt and a minimum of two sketches as input and provide output edits over the neural field faithful to the input conditions.
Meeting all input requirements can be challenging as the text prompt may not match the sketch's semantics, and sketch views may lack coherence.
To undertake this complex task, we conceptually break it down into two subtasks that are easier to handle: one that depends on pure geometric reasoning and the other that exploits the rich semantic knowledge of the generative model. These two subtasks work together, with the former providing a coarse estimate of location and boundary, and the latter adding and refining geometric and texture details through fine-grained operations.

Our experiments highlight the effectiveness of our approach for editing various pretrained NeRF instances. We introduce assorted accessories, objects, and artifacts, which are generated and blended into the original neural field seamlessly.
Finally, we validate our method through quantitative evaluations and ablation studies to assert the contribution of individual components in our method. 
\section{Related work}
\label{sec:related}

\paragraph{Sketch-Based 3D Modeling.}

Since its inception in the late twenties, traditional 2D animation has been concerned with creating believable depictions of 3D forms. Highly acclaimed art guidebook, \textit{The Illusion of Life}, \cite{johnston1981disney}, advocated for solid three-dimensional drawings to practice "weight, depth and balance." With the advancement of computer animation, these principles have been widely adopted
\cite{lasseter1987traditional}.
Sketch-based modeling is typically concerned with stitching and inflating, and user-drawn sketches to 3D meshes \cite{williams1991shading2d}. Starting with Teddy~\cite{igarashi1999teddy}, early works focused on converting scribbles of 2D contours to intermediate forms such as closed polylines \cite{karpenko2006smoothsketch} or implicit functions  \cite{karpenko2002freeform, tai2004prototype, alexe2004interactive, schimdt2005shapeshop, bernhardt2008matisse}. Since lifting a single-view sketch to 3D is an under-constrained problem, additional constraints are usually introduced, such as correlating the inflated thickness with chordal axis depth of curved shapes \cite{igarashi1999teddy}, inferring shape and depth from user annotations \cite{schimdt2005shapeshop, karpenko2006smoothsketch, gingold2009structured, olsen2012natureasketch, tuan2015shading, yeh2017InteractiveHR, li2017bendsketch, jayaraman2018globallyconsistent}, using existing reference models like human figures \cite{turquin2007cloth}, and solving a system based on user constraints \cite{nealen2007fibermesh, joshi2008repousse, sykora2014inkandray, feng2016interactive, dvoro2020monstermash}. 
More recently, data-driven approaches were suggested to lift and reconstruct objects from multi-view sketches with Conv-nets \cite{lun2017sketchreconstruction, delanoy20183d, li2018robust}. 
Our work departs from the former line of research by limiting a generative model to operate within the boundaries of a sketched region. By utilizing the strength of pretrained diffusion models conditioned on language, we avoid the intricacies of explicitly tuning inflation parameters or collecting large-scale train sets while being able to predict texture and shading simultaneously.
\vspace{-15pt}
\paragraph{Diffusion Models.}

Diffusion models \cite{sohl2015diffusion, ho2020ddpm, song2020denoising, song2020improvedsd} have emerged as an increasingly popular technique for generating diverse, high-quality images.
More recently, they've been used to form state-of-the-art text-to-image models \cite{rombach2021highresolution} 
 by introducing language embeddings trained on massive amounts of data \cite{imagen2022saharia, balaji2022eDiff-I}. 
%
Diffusion models are amicable for other conditioning modalities. Related to our work, \cite{meng2022sdedit} introduced conditioning on sketches. \cite{voynov2022sketch} trained a differentiable edge detector used to compute the edge loss per diffusion step, and \cite{cheng2023wacv} allowed finer granularity of control of generated images by distinguishing between sketch, stroke and level of realism.
\cite{zhang2023controlnet} is a contemporary publication which enables additional input conditions by augmenting large-diffusion models on small task-specific datasets.

Another line of works experimented with applying diffusion directly in 3D for generating point clouds \cite{nichol2022pointe} or 2D projections forming the feature representation of neural fields
\cite{shue2022triplanediffusion, anciukevicius2022renderdiffusion}. 
While showing potential, the difficulty lies in scaling them due to the large amount of 3D data required.

\vspace{-15pt}
\paragraph{Neural Fields.}

Neural Radiance Fields (NeRF) \cite{mildenhall2020nerf} have generated massive interest as means of representing 3D scenes using deep neural networks.
Since then, a flurry of works has improved various aspects of optimized neural fields, yielding higher reconstruction quality \cite{barron2021mipnerf, wang2021nerfsr, barron2022mipnerf360, verbin2022refnerf}.
Neural field backbones, in particular, have become more structured and compressed \cite{peng2020convoccnet, takikawa2021nglod, liu2020nsvf, chan2021eg3d, takikawa2022vqad}. The pivotal work of \cite{mueller2022instant} introduced an efficient hash-based representation that allows NeRF optimizations to converge within seconds, effectively paving the way for interactive research directions on neural radiance fields.
%
Recent works have explored interactive editing of neural radiance fields through manipulation of appearance latents
\cite{liu2021editing, niemeyer2021giraffe, park2021hypernerf, chan2021eg3d}, by interacting with proxy representations \cite{yuan2022nerf, bao2022neumesh}, through segmented regions and masks \cite{kobayashi2022distilledfeaturefields, li2022designer3d, mirzaei2022laterf, kuang2022palettenerf} and text-based stylization \cite{gu2021stylenerf, wang2022nerfart, zhang2022arf, mirzaei2022laterf, gao2022get3d}.

Neural fields are an intriguing way to fully generate 3D models because, unlike meshes, they don't depend on topological properties such as genus and subdivision \cite{khalid2022clipmesh, gao2022get3d}.
%
Initial generative text-to-3D attempts with NeRF backbones took advantage of a robust language model \cite{radford2021clip} to align each rendered view on some textual condition \cite{jain2021dreamfields, wang2021clip}. However, without a geometric prior, \cite{radford2021clip} failed to produce multiview-consistent geometry.
\cite{xu2022dream3d} used language guidance to generate 3D shapes based on shape embeddings, but their approach still requires large 3D datasets to generate template geometry.

DreamFusion~\cite{poole2022dreamfusion} and \cite{wang2022scorejacobian} avoid the scarcity of 3D data by harnessing 2D diffusion models pretrained on large-scale, diverse datasets \cite{schuhmann2022laionb}.
Their idea optimizes NeRF representations with score function losses from pretrained diffusion-models \cite{rombach2021highresolution, imagen2022saharia}, where the 2D diffusion process provides gradients for neural fields rendered from random views, and the process is amortized on many different views until an object is formed.
Magic3D~\cite{lin2022magic3d} further improved the quality and performance of generated 3D shapes with a 2-step pipeline which fine-tunes an extracted mesh. Their pipeline also allows for global prompt-based editing and stylization.
Concurrently, Latent-NeRF~\cite{metzer2022latent} suggested optimizing neural fields in the diffusion latent space. Their work also suggested 3D bounded volumes as an additional constraint for guiding the generation process.

Our work builds on a simplified framework of \cite{poole2022dreamfusion}, which operates in color space and aims to combine traditional sketch-based modeling constraints with the generative power of recent advances in the field. Our pipeline is a zero-shot generative setting, requiring no dataset and only text and sketch inputs from the user.

\section{Method}
\label{sec:method}

\begin{figure*}[t]
\centering
  \includegraphics[width=0.9\linewidth]{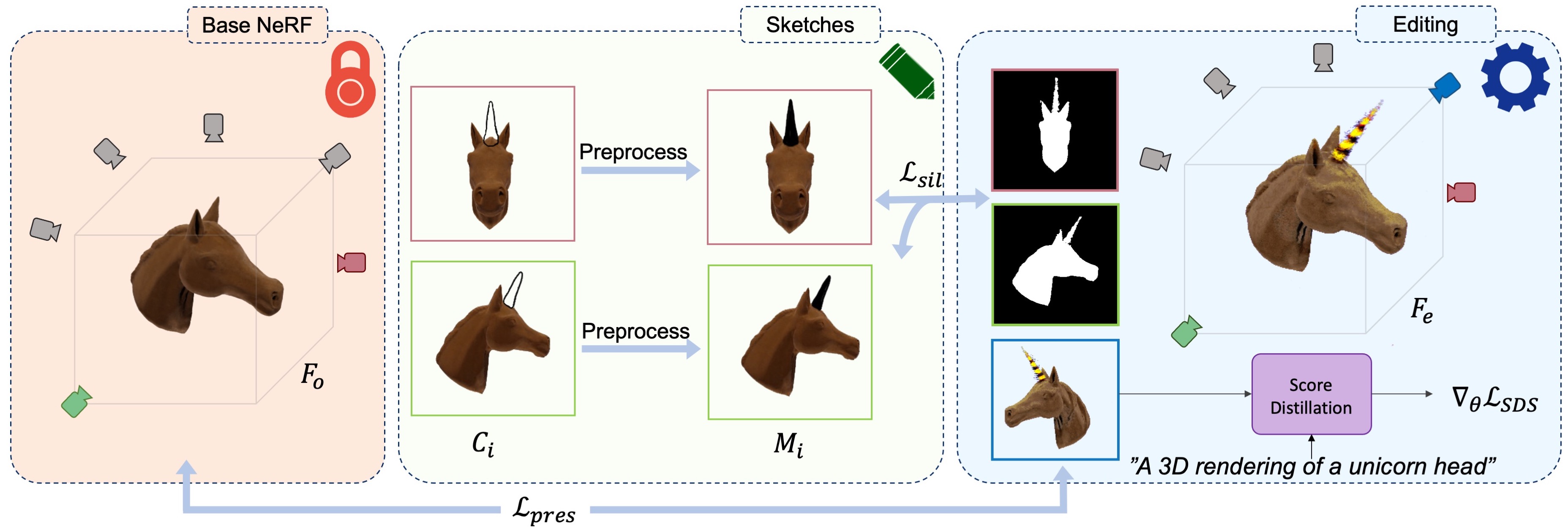}

  \captionsetup{font=small}
  \caption{An overview of SKED. We render the base NeRF model $\basenerf$ from at least two views and sketch over them ($C_i$). The input to the editing algorithm is these sketches preprocessed to masks ($M_i$) and a text prompt. In each iteration similar to DreamFusion \cite{poole2022dreamfusion}, we render a random view and apply the Score Distillation Loss to semantically align with the text prompt. Additionally, we compute $\mathcal{L}_{pres}$ to preserve the base NeRF by constraining $\edittednerf$'s density and color output to be similar to $\basenerf$ away from the sketch regions. Finally, we use the object mask renderings of the sketch views to define $\mathcal{L}_{sil}$. This loss ensures that the object mask occupies the sketch regions.}
  \label{fig:overview}
\end{figure*}

In this section, we present our approach to performing text-based NeRF editing, controlled by a few given sketches. Our approach to addressing this issue involves dividing the problem into two substantially easier tasks. First, a rough 3D space that necessitates adjustment is defined using the provided sketches, which helps in guiding the geometry modifications.
Second, we use the score distillation sampling method~\cite{poole2022dreamfusion} on a text-to-image latent diffusion model to generate fine-detailed and realistic edits based on the text prompt given to the model (see Fig.~\ref{fig:overview}).

To produce meaningful edits that adhere to the sketches, we design two novel objective functions: one to preserve the original density and radiance fields, and the second to alter the added mass in a way that respects the given sketches. In the following, we describe our loss functions and provide details on how to apply sketch-based text-guided editing. We include a background on Latent diffusion models~\cite{rombach2021highresolution} and Score distillation sampling (SDS)~\cite{poole2022dreamfusion} which our method is built upon, in the supplementary material.

\subsection{SKED}
\label{subsec:sked}
\begin{figure}[t]
    \centering
    \includegraphics[width=\linewidth]{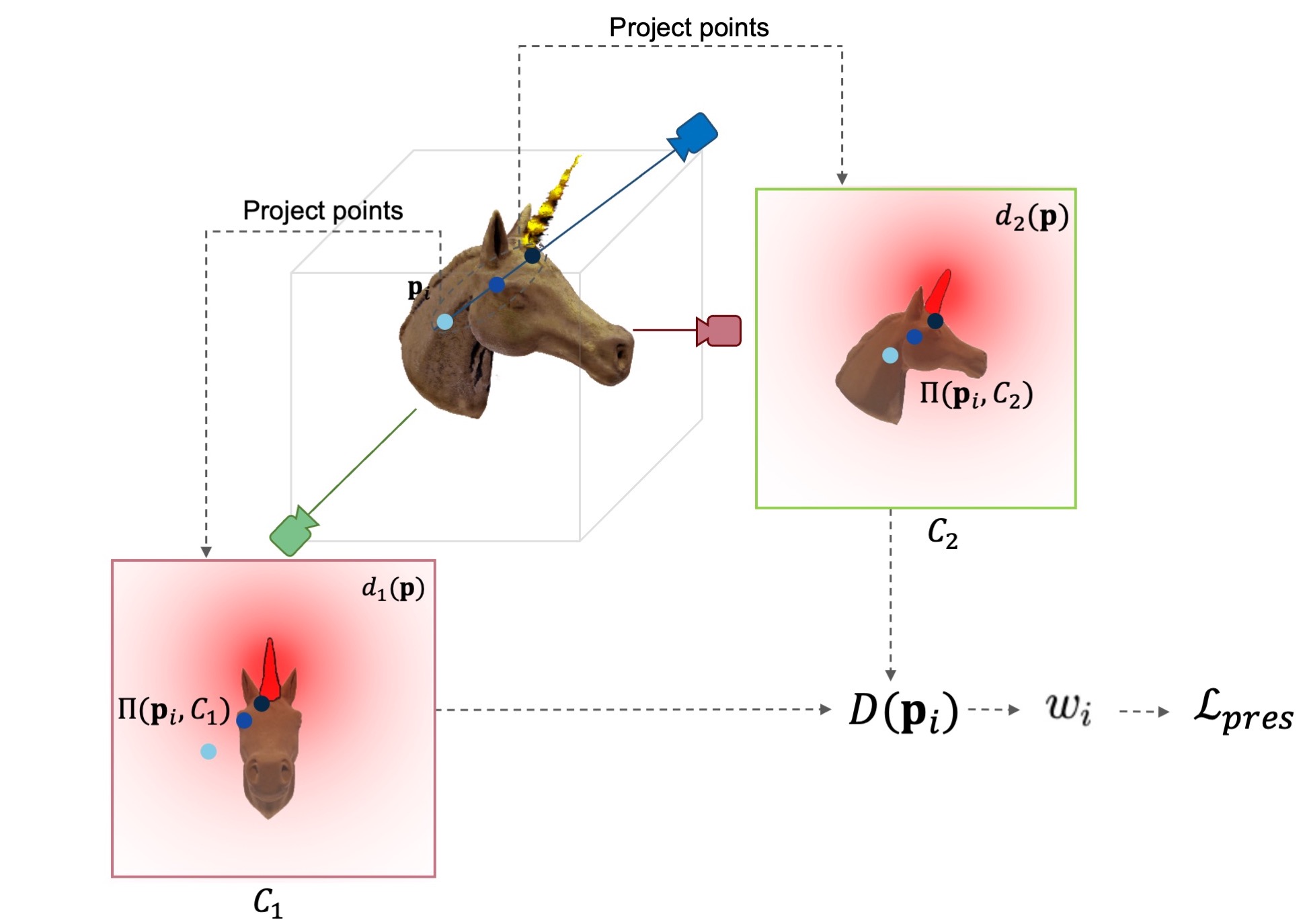}
    \captionsetup{font=small}
    \caption{3D points $\coord$ sampled at random views are projected to the sketch views $C_j$ to obtain $\Pi(\coord, C_j)$. In each $C_j$, distance $d_j$, between projected points and the pixels containing the sketch masks is computed. The red color in $C_1$ and $C_2$ demonstrates $d_1(\textbf{p})$ and $d_2(\textbf{p})$ in image space respectively. Finally, for each 3D point, $d_j(\coord)$s are averaged to get the distance $D(\coord)$ to all sketch views. $D(\coord)$ is used as the weights of the points $w_i$ in $\mathcal{L}_{pres}$. }
    \label{fig:fig3}
\end{figure}

As demonstrated in Fig.~\ref{fig:overview}, the starting point of our algorithm is a base NeRF model, $\basenerfFull$, which maps 3D coordinates and unit rays to color and volumetric density. $\basenerf$ is obtained through either reconstruction from multiview images \cite{mueller2022instant}, or a text-to-3D pipline \cite{poole2022dreamfusion}. One can use $\basenerf$ to render multiple views of the neural field and sketch over them to specify spatial cues of their desired edits. Let $\{C\}_{i=1}^N$ be renderings of $\basenerf$ from $N$ different views, on which we have drawn sketches. These sketches could be masks specifying the region of edit, or closed curves specifying the outer edges of the region of edit. Either way, the input to our algorithm would be preprocessed to masks $\{M\}_{i=1}^N$ where $M_i = \{m_1, m_2, ..., m_S\}$ are a set of pixels which are inside the sketch region. We call these renderings \emph{sketch canvases}, and the views they were rendered from \emph{sketch views}. Additionally, the algorithm takes as input a text prompt $T$ which defines the semantics of the edit. Similar to DreamFusion, our method is an iterative algorithm. We begin by initializing an editable copy of the base neural field, $\edittednerf=\basenerf$. At each iteration, we sample a random view and use $\edittednerf$ to render the 3D object from that view. We use the rendered image to calculate the gradient of the SDS loss and push $\edittednerf$ to the high-density regions of the probability distribution function $\mathbb{P}(\edittednerf | T)$ i.e. a NeRF model which its underlying 3D object adheres to text $T$. However, through our experiments, we found that simply using this process with only text as guidance would drastically change the original field outside of the sketched region. Therefore through two novel loss functions, we attempt to control the editing process in a way that the final output coresponds to the sketches is semantically meaningful and is faithful to the base neural field.
\vspace{-15pt}
\paragraph{Preservation Loss:} One of the main criteria of a good 3D editing algorithm is that the geometry and color of the base object are preserved through the editing process. We encourage this by utilizing an objective we call the preservation loss $\mathcal{L}_{pres}$. At each iteration of the algorithm, we render an image with $\edittednerf$ from a random camera viewpoint. We modify the raymarching algorithm of NeRF such that when sampling points $\coord \in {\reals}^{3}$, to query $\edittednerf$ for density and color values, we also compute a per-coordinate sketch-weight denoted as $\preservationweight$.
The key idea of our algorithm is that for each point $\coord$, we decide whether it should be changed by calculating a distance from the point to the sketch masks. We aim to modify the density and radiance only when $\coord$ is in the proximity of the sketched regions while retaining the original density and radiance for points that are far from the sketched region.
Therefore, we first need to define a method for computing the distance of a 3D point to multiview $2D$ sketches. We do so by projecting $\coord$ to each of the sketch views and computing a per-view distance of projected points to sketch regions as:
\begin{equation}
    d_j(\coord) = \min_k{|| \left \lfloor \Pi(\coord, C_j) + \frac{1}{2} \right \rfloor - m_k|| ^ 2},
    \label{eq:per_view_dist}
\end{equation}
where $d_j(\coord)$ is the per-view distance function and $\Pi(\coord, C_j)$ is the projection of 3D point $p_i$ to sketch view $C_j$, rounded to the nearest integer (Fig.~\ref{fig:fig3}). This expression computes the minimum distance of the projection of point $p_i$ to the sketch regions in view $j$. By taking the mean of all the per-view distances, we can define the distance of point $p_i$ to the multiview sketches $D(\coord)~=~\frac{1}{N}\sum_{j=1}^N{d_j(\coord)}$. We use the mean of distances, as it relaxes the constraint when sketches are not fully aligned, and introduces additional smoothness to the function.

Now that we have established the distance function, we can define $\mathcal{L}_{pres}$. Using $\basenerf$, the base NeRF instance with frozen parameters, we define $\mathcal{L}_{pres}$ as:

\begin{equation}
    \begin{split}
          \mathcal{L}_{pres} = \frac{1}{K}\sum_{i=1}^K w_i [CE(\outeditalpha, \overbar{\outbasealpha})
    + \lambda_{c}\overbar{\outbasealpha}||\outeditcolor - \outbasecolor|| ^ 2],
    \end{split}
\end{equation}
where $\overbar{\outbasealpha}$ is the occupancy of the base object derived from $\basenerf$ by thresholding the ground truth density of $\basenerf(\coord ; \theta)$. Following \cite{mildenhall2020nerf}, we have $\outeditalpha=1-exp(-\outeditdensity\delta)$, such that $\outeditdensity$ is the edited neural field density $\edittednerf(\coord ; \phi)$, and $\delta$ is the step distance between samples along a ray. $CE$ is the cross entropy loss. Furthermore, $\outeditcolor$ and $\outbasecolor$ are the color and ground truth color of $\coord$ derived from $\edittednerf$ and $\basenerf$ respectively, and $\lambda_{c}$ is a hyperparameter controlling the importance of color preservation in the editing process. We limit color preservation to the occupied region of $\basenerf$. Tightly constraining the color may drive the model to diverge, hence $\lambda_{c}$ is chosen such that density preservation takes higher priority.
The preservation strength of each coordinate is modulated by: 
\begin{equation}
    \preservationweight = 1 - \exp(-\frac{D(\coord) ^ 2}{2\sensitivity^2}).
    \label{eq:modulate}
\end{equation}
Intuitively, $\preservationweight$ controls the importance of the loss for each point based on the distance $D(\coord)$. The sensitivity of $\preservationweight$ to $D(\coord)$ is controlled by the hyperparameter $\sensitivity$, where lower values tighten the constraint on the optimization such that only the sketch region is modified. Higher values allow for a softer falloff region, which allows the optimized volume to better blend with the base model.

\vspace{-15pt}
\paragraph{Silhouette Loss:} Another essential criterion is to respect the sketched regions, i.e. the new density mass added to $\edittednerf$ should occupy the regions specified by the sketches. We enforce this by rendering the object masks of all sketch views. We then maximize the values of the object masks in the sketched regions by minimizing the following loss: 

\begin{equation}
    \mathcal{L}_{sil} = \frac{1}{H.W.N}\sum_{j=1}^{N} \sum_{i=1}^{H. W} -\mathbb{I}_{M_{j}}(\pixel) \log C_{j}^{\alpha}(\pixel).
\end{equation}
In this equation, $H$ and $W$ are the dimensions of the rendered object masks, $\mathbb{I}_{M_{j}}$ is an indicator function that is equal to 1 if pixel $\pixel \in \reals^2$ is in a sketched region and 0 otherwise and $C_{j}^{\alpha}$ is the alpha object mask rendered with $\edittednerfFull$ from each sketch view.

\vspace{-10pt}
\paragraph{Optimization:} Similar to prior generative works on NeRFs ~\cite{ jain2021dreamfields, lin2022magic3d, poole2022dreamfusion} we use an additional objective $\mathcal{L}_{sp}$ to enforce sparsity of the object by minimizing the entropy of the object masks in each view. Therefore the final objective of our editing process is:
\begin{equation}
\begin{split}
        \mathcal{L}_{total} = \mathcal{L}_{SDS} + \lambda_{pres}\mathcal{L}_{pres} + \lambda_{sil}\mathcal{L}_{sil} + \lambda_{sp}\mathcal{L}_{sp},
\end{split}
\end{equation}
where $\lambda_{pres}, \lambda_{sil}$ and $\lambda_{sp}$ are the weights of the different terms in our objective.

We use Instant-NGP~\cite{mueller2022instant} as our neural renderer for its performance and memory efficiency. To avoid sampling empty spaces, this framework keeps an occupancy grid for tracking empty regions. The occupancy grid is used during raymarching to efficiently skip samples in empty spaces. In addition, the grid is periodically pruned during training to keep it aligned with the hashed feature structure. In our editing process, if the occupancy grid of $\basenerf$ is used without change, the model may initially avoid sampling points in the sketch regions, preventing correct gradient flow and forcing our framework to rely on random alterations of the occupancy grid.

To alleviate this problem, we find the bounding boxes of the sketch masks $M$ and intersect them in $\reals^3$ to define a coarse editing region in 3D space. We manually turn on the occupancy grid bits of $\edittednerf$ within the sketch intersection region. In addition, we define a warm-up period at the beginning of optimization, where we avoid pruning the occupancy grid to help the model solidify the edited region, and prevent it from culling it as empty space.

\section{Results and Evaluation}

\label{sec:results}
\begin{table*}
\centering

\captionsetup{font=small}
\caption{
Fidelity of base field. To assess a method's ability to preserve the original content, we measure the \textbf{PSNR $\uparrow$} of the method's output against renderings from the base model. \ours  \enspace \textit{(no-preserve)} refers to a variant of our method which doesn't apply $\mathcal{L}_{pres}$. Text-Only refers to a public re-implementation of \cite{poole2022dreamfusion}.
}
\label{tab:psnr}

\begin{tabular}{l \tsm \tsm c \ts c \tsm c \ts c \tsm c \ts c \tsm c \ts c \tsm c \ts c \tsm c}
\toprule

{Method} & 
\multicolumn{2}{c \tsm  \ts}{{Cat}} &
\multicolumn{2}{c \tsm}{{Cupcake}} &
\multicolumn{2}{c \tsm}{{Horse}} &
\multicolumn{2}{c \tsm}{{Sundae}} &
\multicolumn{2}{c \tsm}{{Plant}} &
{{Mean}} \\

& 
\multicolumn{2}{c \tsm  \ts}{{\textit{+chef hat}}} &
\multicolumn{2}{c \tsm}{{\textit{+candle}}} &
\multicolumn{2}{c \tsm}{{\textit{+horn}}} &
\multicolumn{2}{c \tsm}{{\textit{+cherry}}} &
\multicolumn{2}{c \tsm}{{\textit{+flower}}} &
\\

& View 1 & View 2
& View 1 & View 2
& View 1 & View 2
& View 1 & View 2
& View 1 & View 2
&
\\
\midrule

\ours  & 
\textbf{31.05} & \textbf{34.13} & 
\textbf{23.73} & \textbf{25.98} & 
\textbf{32.45} & \textbf{31.46} & 
\textbf{26.47} & \textbf{25.99} & 
\textbf{21.71} & \textbf{22.31} & 
\textbf{27.53}
\\

\ours \enspace \textit{(no-preserve)}  & 
15.58 & 16.59 & 
20.12 & 19.47& 
18.02 & 16.52 & 
17.39 & 17.44 & 
10.16 & 10.12 & 
16.14
\\

Text-Only \cite{stable-dreamfusion}  & 
15.63 & 16.78 & 
17.38 & 17.15 & 
16.69 & 15.09 & 
20.57 & 20.74 & 
13.75 & 12.68 & 
16.65
\\

\bottomrule

\end{tabular}
\end{table*}

\subsection{Implementation details}
We use Stable-DreamFusion's open-source GitHub repository~\cite{stable-dreamfusion} and integrate with kaolin-wisp's renderer for an interactive UI~\cite{KaolinWispLibrary}.
In all our experiments, unless stated otherwise, we set $\lambda_{pres}, \lambda_{sil}$, $\lambda_{sp}$ and $\lambda_{c}$ to $5\times 10^{-6}$, $1$, $5\times 10^{-4}$ and $5$ respectively.
The guidance scale for classifier free guidance in the Stable Diffusion model is set to 100 and timesteps for noise scheduling are uniformly sampled in the range of (20, 980) in each iteration.
The warm-up period for the occupancy grid pruning is set to 1,000 iterations. We also set the camera pose range to ensure that the sketch region remains visible in all sampled views. For large sketch regions, we use a Kd-tree 
\cite{2020SciPy-NMeth}
to implement Equation
3
with efficient queries. We use the ADAM
~\cite{kingma2014adam}
optimizer with a learning rate of $0.005$ and apply the exponential scheduler with a decay of $0.1$ by the end of the optimization.  We run our algorithm for $10,000$ iterations,  taking approximately 30-40 minutes on a single NVIDIA RTX 3090 GPU.
For our experiments, we use both publicly available 3D assets, and artificial assets generated by Stable-DreamFusion~\cite{stable-dreamfusion} guided only by text. We use the v1.4 version of the Stable diffusion~\cite{rombach2021highresolution} model. Unless specified otherwise, we use the same default hyperparameters settings throughout all experiments depicted in the paper.

\begin{figure*}[t]
  \centering\includegraphics[width=0.95\linewidth]{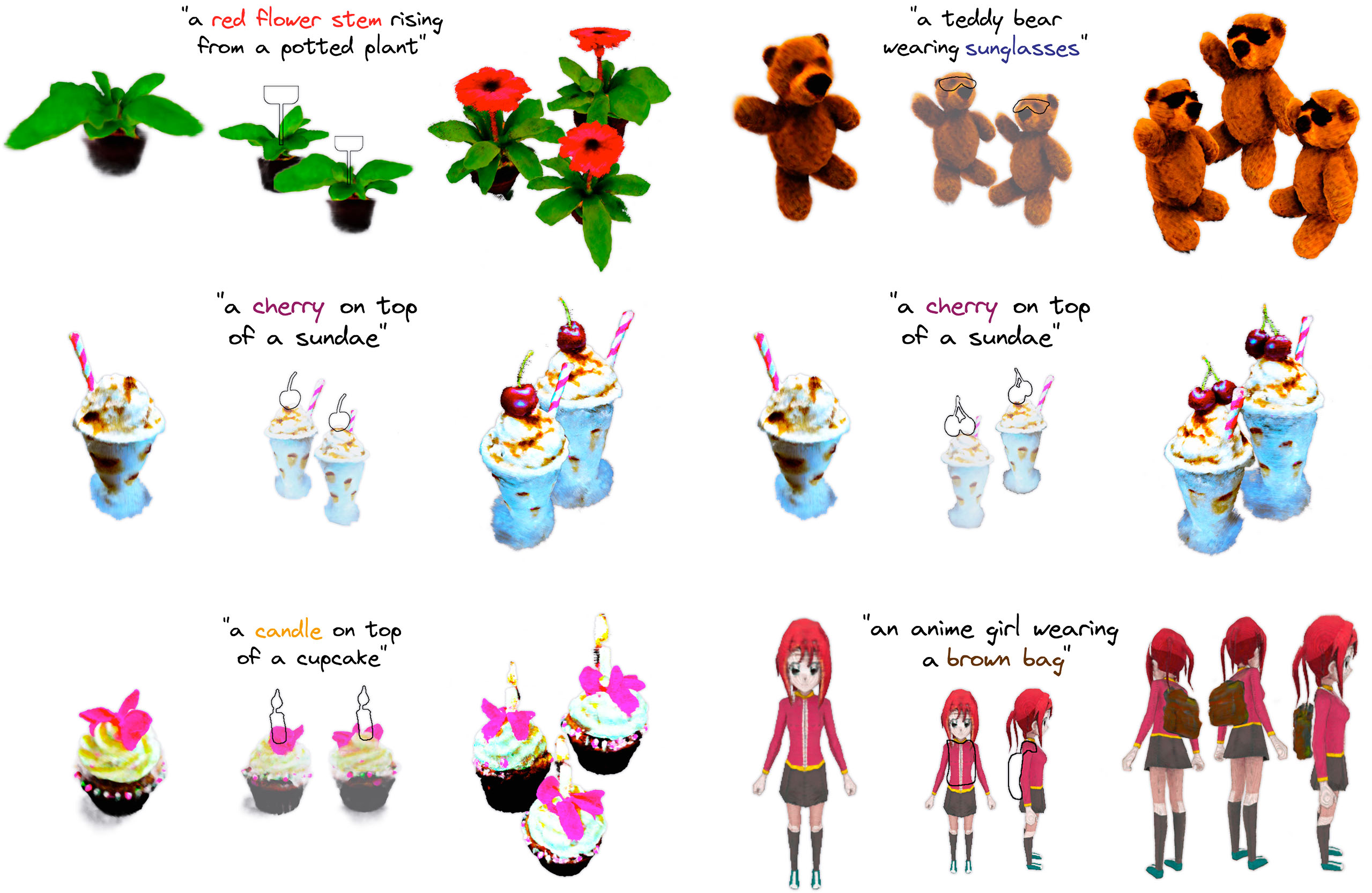}
  \captionsetup{font=small}
  \caption{Examples of using SKED to edit various objects reconstructed with InstantNGP~\cite{mueller2022instant} (anime girl) or generated with DreamFusion~\cite{poole2022dreamfusion} (plant, sundae, teddy bear, sundae, cupcake). All examples were edited using two sketch views and the text prompt.}
  \label{fig:results}
\end{figure*}

\begin{figure*}[t]
    \centering
    \begin{overpic}[width=0.9\textwidth,tics=10, trim=0 0 0 0,clip]{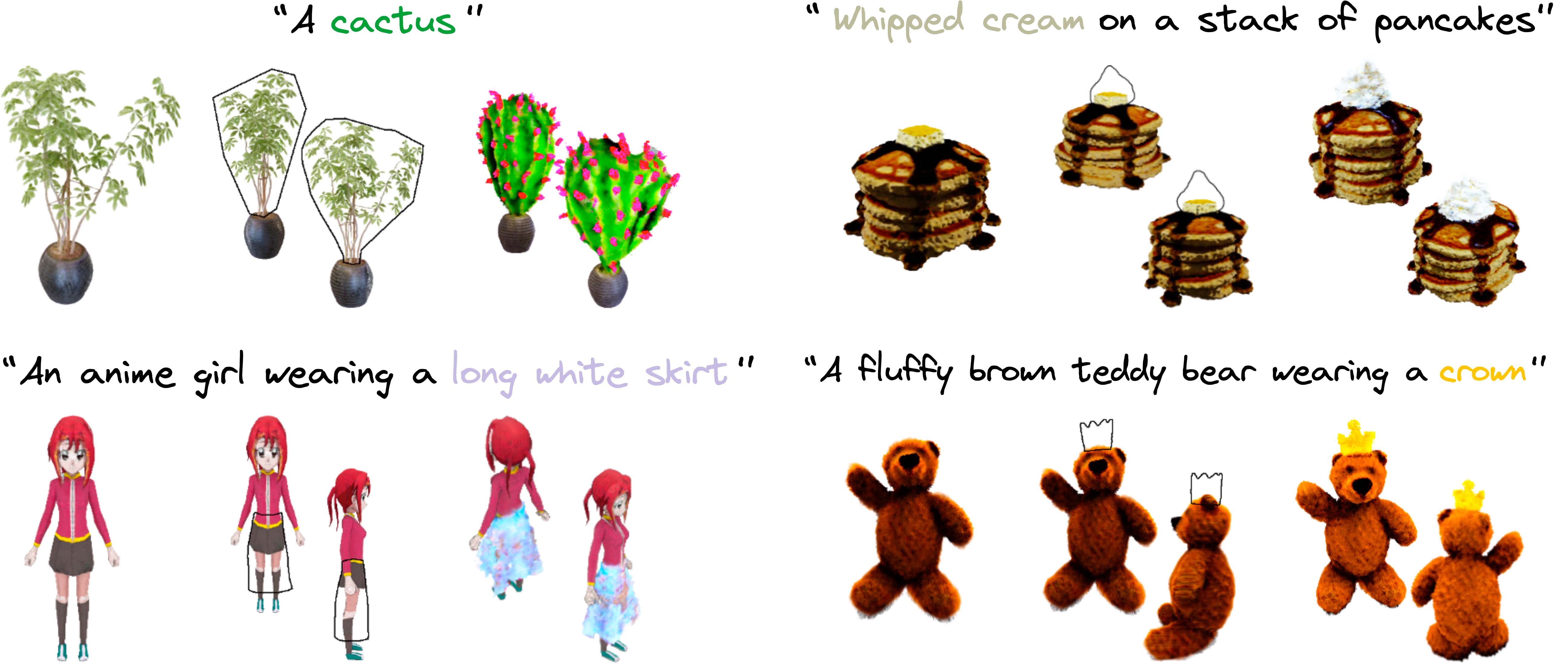}
    \end{overpic}
    \captionsetup{font=small}
    \caption{Various types of edits. \ours{} is capable of overwriting parts the base model (Cactus, Skirt), as well as adding new details (Pancake, Teddy).}
    \label{fig:additional}
\end{figure*}

\subsection{Qualitative Results}
\label{subsec:qualitative}

\textbf{Sketch and Text Control.} Fig.~\ref{fig:results} demonstrates examples 
of SKED on a variety of objects and shapes. Evidently, our method is able to satisfy the coarse geometry defined by user sketches, and at the same time naturally blend semantic details according to the given text-prompts. Note that the sketch is not required to be accurate or tight: by making the contour curve more complex, the user can further force the pipeline to generate a specific shape.

\begin{figure*}[t]
  \centering
  \includegraphics[width=0.99\linewidth]{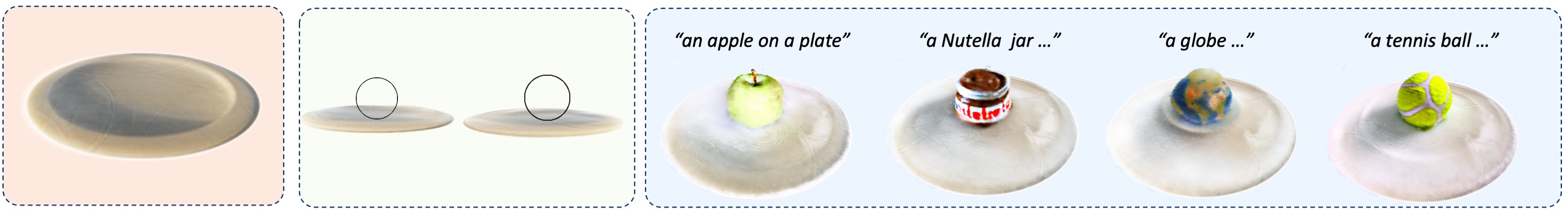}
  \captionsetup{font=small}
  \caption{Examples with a single set of sketches and a variety of text prompts. Our method is able to respect the geometry of the sketches while adding details to fit the different prompts' semantics.}
  \label{fig:balls}
\end{figure*}

Next, we show that given a fixed pair of multiview sketches, our method produces semantic details to fit a diverse set of text-prompts (Fig.~\ref{fig:balls}). 
Note that our method can generate details within the sketch boundary (e.g. Nutella jar) even if the sketch doesn't match the text-prompt description.
In Fig.~\ref{fig:results} we also present the complementary case, by re-using the same text prompts and switching through different sketch sets, our method has the flexibility to produce localized edits (i.e., "cherry on top of a sundae").

Additionally, In Fig.~\ref{fig:additional} we demonstrate the ability of our method in performing various types of edits. We are able to perform both additive edits (crown on teddy's head or whipped cream on pancake) and replacement edits where we overwrite a part of the object with a different part (tree to cactus or the long white skirt).

\begin{wrapfigure}{r}{0.19\textwidth} 
\vspace{-2pt}
  \begin{center}
    \includegraphics[width=0.19\textwidth]{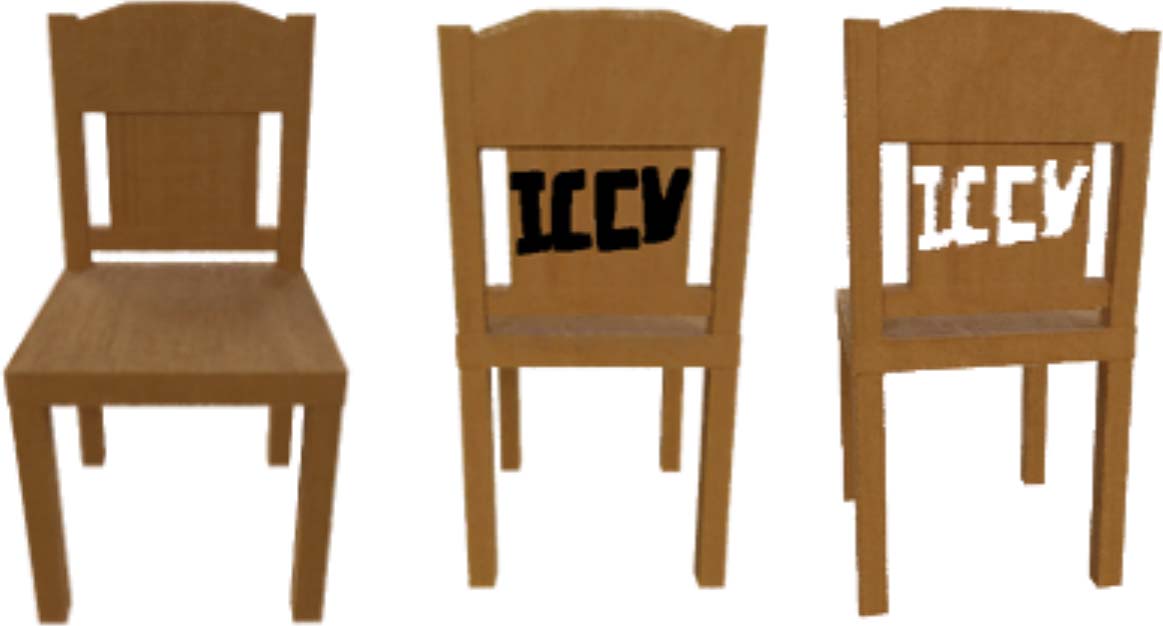}
  \end{center}
  \vspace{-15pt}
  \vspace{1pt}
\end{wrapfigure} 
Although not text based, we are also able to perform simple carving edits by masking off the 3D grid in the sketched regions (inset).

\begin{figure}[t]
    \centering
    \includegraphics[width=0.85\linewidth]{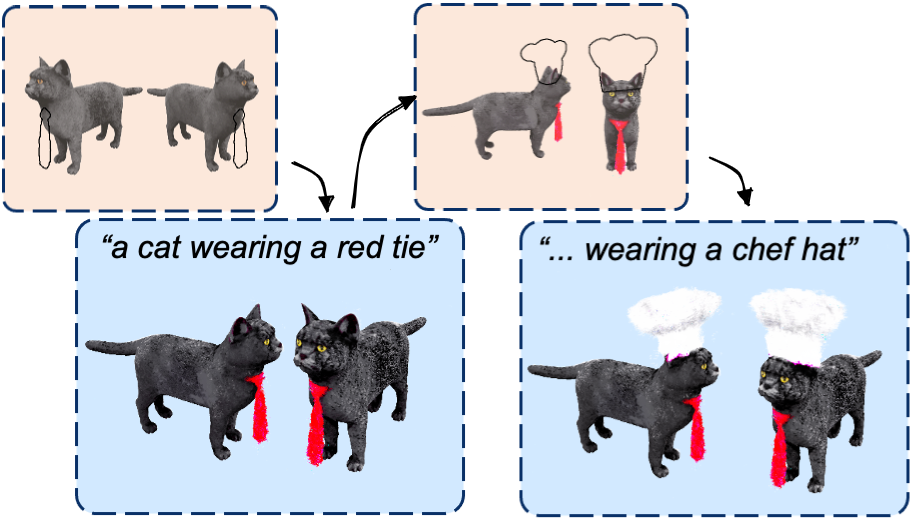}
    \captionsetup{font=small}
    \caption{Progressive editing. The cat is first edited by adding a red tie, and then a chef hat is added in a subsequent edit.}
    \label{fig:progressive}
\end{figure}

\begin{figure}
   \centering
    \begin{overpic}[width=\linewidth,tics=10, trim=0 0 0 0,clip]{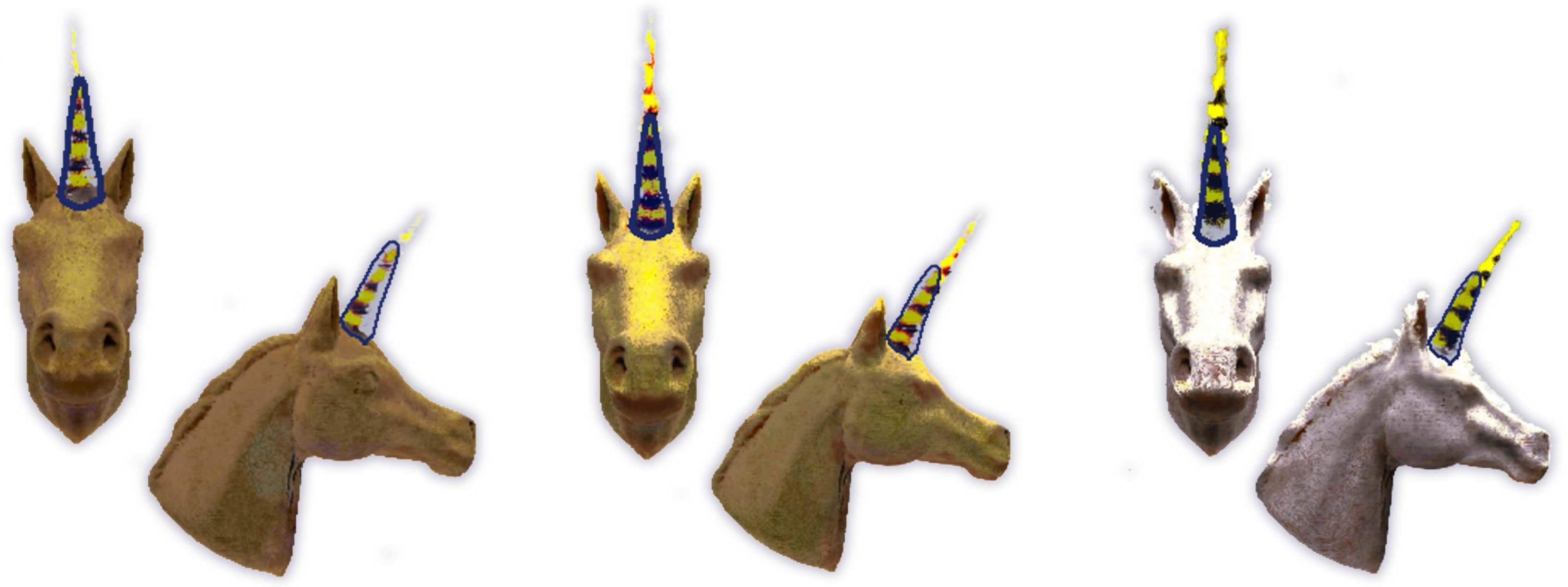}
    \put(10,-3){$\beta=0.005$}
    \put(45,-3){$\beta=0.1$}
    \put(78.0,-3){$\beta=1.0$}
   \end{overpic}
   \captionsetup{font=small}
    \caption{Sensitivity control. Depending on the sensitivity value determined by $\beta$ in Eq.~\ref{eq:modulate}, our method can either edit only the sketched region and minimally modify the rest of the neural field, or produce larger edits outside the sketch regions (softer blending). We display the overlay of sketches v.s. edited output.}
    \label{fig:overlay}
 \end{figure}

\textbf{Base Model Distribution.} \label{subsubsec:basemodeldist} Our method assumes edits are applied on top of a base NeRF model. We explore both pretrained reconstructions from multiview images, generated with \cite{mueller2022instant}, and generated outputs from DreamFusion~\cite{poole2022dreamfusion} using the same diffusion model we use for editing \cite{rombach2021highresolution}. 
The renderings of reconstructed objects are assumed to follow the distribution of the underlying diffusion model.
Our method performs successfully on both reconstructed examples: Anime Girl, Plate, Cat, Horse (Fig. \ref{fig:results}, \ref{fig:balls}, \ref{fig:progressive}, \ref{fig:overlay}) and generated ones: Plant, Teddy, Cupcake, Sundae (Fig.~\ref{fig:results}).

\begin{table*}
\centering

\captionsetup{font=small}
\caption{Sketch alignment score. We measure the similarity between the user input and generated result by Intersection-over-Sketch (\textbf{IoS $\uparrow$}). The IoS is calculated using the intersection between two views of filled sketches, $M_i$, and the alpha mask of generated edit $C_i^\alpha$. See Section~\ref{subsec:quantitative} for elaborate details of this metric. The SKED \textit{(no-silh)} variant, which runs with $\mathcal{L}_{pres}$ and without $\mathcal{L}_{sil}$ avoids generating content in the sketch region (see also Fig.~\ref{fig:ablations})
}
\label{tab:ios}

\begin{tabular}{l \tsm c \ts c \tsm c \ts c \tsm c \ts c \tsm c \ts c \tsm c \ts c \tsm c}
\toprule

{Method} & 
\multicolumn{2}{c \tsm  \ts}{{Cat}} &
\multicolumn{2}{c \tsm}{{Cupcake}} &
\multicolumn{2}{c \tsm}{{Horse}} &
\multicolumn{2}{c \tsm}{{Sundae}} &
\multicolumn{2}{c \tsm}{{Plant}} &
{{Mean}} \\

& 
\multicolumn{2}{c \tsm  \ts}{{\textit{+chef hat}}} &
\multicolumn{2}{c \tsm}{{\textit{+candle}}} &
\multicolumn{2}{c \tsm}{{\textit{+horn}}} &
\multicolumn{2}{c \tsm}{{\textit{+cherry}}} &
\multicolumn{2}{c \tsm}{{\textit{+flower}}} &
\\

& View 1 & View 2
& View 1 & View 2
& View 1 & View 2
& View 1 & View 2
& View 1 & View 2
&
\\
\midrule

\ours  & 
\textbf{0.9384} & \textbf{0.9689} & 
\textbf{0.8364} & \textbf{0.8875} & 
\textbf{0.6423} & \textbf{0.5363} & 
\textbf{0.7817} & \textbf{0.9096} & 
\textbf{0.9388} & \textbf{0.7801} & 
\textbf{0.8220} 
\\


SKED \textit{(no-silh)} & 
0.0196 & 0.0176 & 
0.0263 & 0.0209 & 
0.0090 & 0.0077 & 
0.0541 & 0,0506 & 
0.0024 & 0.0028 & 
0.0211 
\\

\bottomrule

\end{tabular}
\end{table*}

\begin{table*}
\centering

\captionsetup{font=small}
\caption{Semantic alignment score. We measure the \textbf{CLIP-similarity} 
\cite{radford2021clip}
$\uparrow$ of the rendered method output with the clip embedding of the input text prompt.  Text-Only refers to a public re-implementation of 
\cite{poole2022dreamfusion}.
The qualitative equivalents of Cat and Plant examples are depicted in Fig.
7 from the main paper:
Compared with 
\cite{poole2022dreamfusion}
 which changes the structure of the base model to satisfy the text semantics, our method preserves the base model, while also maintaining semantic correlation with the text.
}
\label{tab:clipsim}

\begin{tabular}{l \tsm c \tsm c \tsm c \tsm c \tsm c \tsm c}
\toprule

{Method} & {Cat} & {Cupcake} & {Horse} & {Sundae} & {Plant} & {Mean} \\
& {\textit{+chef hat}} & {\textit{+candle}} & {\textit{+horn}} & {\textit{+cherry}} & {\textit{+flower}} & \\
\midrule

\ours  & 
0.2336 $\pm$ 1.1e-3 & 
\textbf{0.2849 $\pm$ 4.2e-3} & 
\textbf{0.2943 $\pm$ 4.8e-3} & 
0.2635 $\pm$ 2.4e-3 & 
\textbf{0.2933 $\pm$ 4.0e-3} & 
0.2739 
\\

Text-Only~\cite{stable-dreamfusion} &
\textbf{0.2744 $\pm$ 4.4e-3} & 
0.2818 $\pm$ 6.2e-3 & 
0.2928 $\pm$ 8.4e-3 & 
\textbf{0.2674 $\pm$ 4.9e-3} & 
0.2865 $\pm$ 3.3e-3 & 
\textbf{0.2806}
\\

\bottomrule

\end{tabular}
\end{table*}

\textbf{Progressive Editing.} Our method can be used as a sequential editor where one reconstructs or generates a NeRF, and then progressively edits it. In Fig.~\ref{fig:progressive} we exhibit a two-step editing by first reconstructing a cat object with \cite{mueller2022instant}, then generating a red tie, followed by adding a chef hat.

\textbf{Preservation Sensitivity.} We also demonstrate the overlay of the sketch masks with the edited NeRF rendered from the sketch views and the effect of changing $\beta$ (Eq.~\ref{eq:modulate}) in Fig.~\ref{fig:overlay}. It is evident that increasing $\beta$ changes the base NeRF more and more edits appear outside the sketch regions.


%

\subsection{Quantitative Results}
\label{subsec:quantitative}
To the best of our knowledge, our method is the first to perform sketch-guided editing on neural fields. Hence, at the absence of an existing benchmark for systematic comparison, we also suggest a series of tests to quantify various aspects of our method. We conduct our evaluation using a set of five representative samples using the setting from Section~\ref{sec:results}. Each sample includes a base shape, a pair of hand drawn sketches and a guiding text prompt. Comparisons to "Text-only" ignore the input sketches and apply prompt editing. For a fair comparison, all experiments use the same diffusion model and implementation framework of~\cite{stable-dreamfusion}. In the following, we establish that all three metrics are necessary to quantify the method's value.

\textbf{Base Model Fidelity.} We quantify our method's ability to preserve the base field outside of the sketch area using PSNR (Table~\ref{tab:psnr}). As ground truth, we use $\{C \backslash M \}_{i=1}^{N=2}$, the renderings from the base model, excluding the filled sketch regions. We measure the PSNR w.r.t. to the output sketch view rendered with edited field $\edittednerf$, and the output from \cite{poole2022dreamfusion} using the same camera view. Our results show that across all inputs, our method consistently preserves the original base field content, compared to the Text-only method which lacks this ability. Note that a method may obtain a perfect PSNR if it does not change the original neural field. Therefore, we further measure the quality of change as well.

\begin{figure*}[t]
    \centering
    \begin{overpic}[width=1\textwidth,tics=10, trim=0 0 0 0,clip]{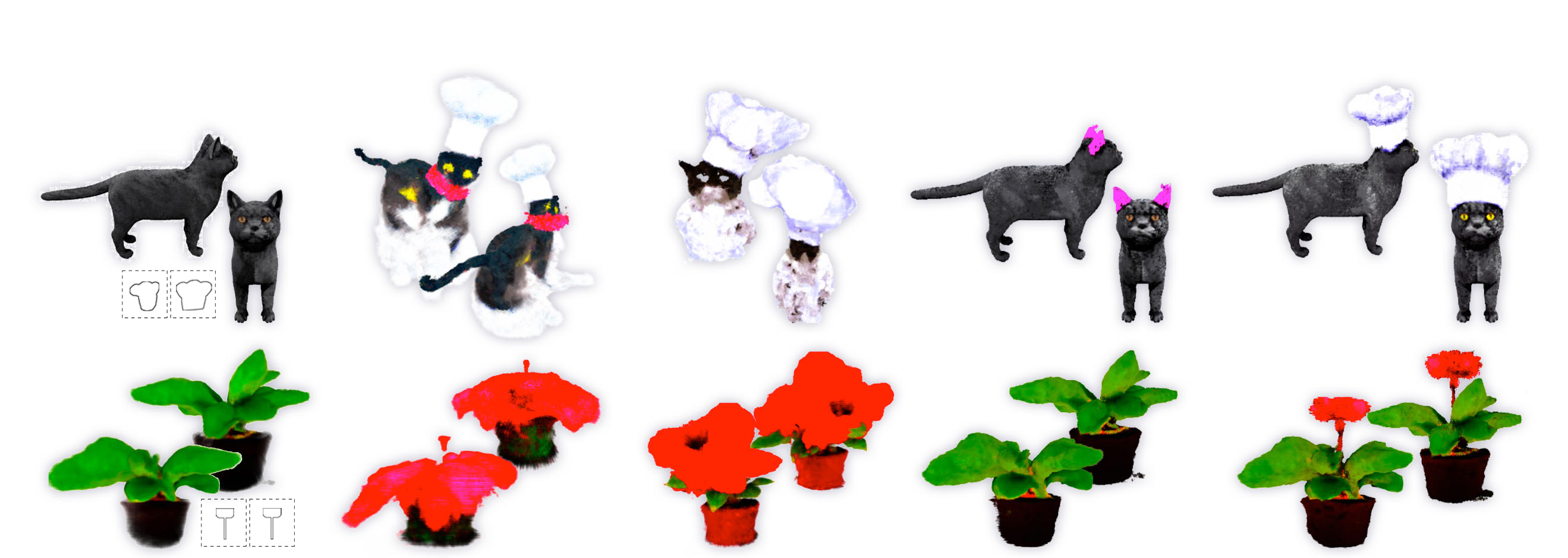}
    \put(6.5,33){Base Model}
    \put(26.0,33){Text Only}
    \put(46.0,34){Ours}
    \put(44,32){$\Lstroke_{pres}, \mathcal{L}_{sil}$}
    \put(64.0,34){Ours}
    \put(62,32){$\mathcal{L}_{pres}, \Lstroke_{sil}$}
    \put(86.25,34){Ours}
    \put(87.0,32){\textit{full}}
    \end{overpic}
    \captionsetup{font=small}
    \caption{Ablation Study. We demonstrate of the effect of our suggested losses on editing neural fields (zoom in for details). The prompts used are \textit{"A cat wearing a \textcolor{Periwinkle}{chef hat}"} and \textit{"A \textcolor{red}{red flower stem} rising from a potted plant"}. All methods were initialized with the same base models (leftmost column), and optimized with the same diffusion model \cite{rombach2021highresolution}. Text-only uses the public re-implementation of \cite{poole2022dreamfusion}. The rightmost method shows our full pipeline, compared to ablated versions of it omitting $\mathcal{L}_{pres}$ and $\mathcal{L}_{sil}$ respectively.}
    \label{fig:ablations}
\end{figure*}

\textbf{Sketch Filling.} To gauge whether our method respects the user sketches, we measure the ratio of sketch area filled with generated mass (Table~\ref{tab:ios}). We denote the metric we use as \textit{Intersection-over-Sketch}, and define it as $IoS ~=~\sum_{i=1}^{N}{{|M_i \cap C_{i}^{\alpha}|}/{|M_i|}}$.
Here, $M_i$ is the sketch region and $C_{i}^{\alpha}$ is the thresholded alpha mask rendered with $\edittednerf$ from each sketch view. To ensure the metric is resilient to alpha thresholding values, the score we report is averaged over 9 thresholding values we apply to $C_i^\alpha$, ranging at $[25, 225]$. We point out that a high $IoS$ score by itself does not guarantee a high quality output, i.e. a method can cheat by simply filling the sketch region with some fixed color.

\textbf{Semantic Alignment.} We assess if our method generates semantically meaningful content aligning with the text-prompt using Clip-similarity~\cite{radford2021clip}. In Table~\ref{tab:clipsim}, we present the evaluations which demonstrate that although we don't optimize for CLIP performance directly, our method achieves comparable results with Text-Only. We perform this experiment by sampling forty views around each edited object and averaging the CLIP similarity of each view to the corresponding texts.


\subsection{Ablation Studies}

 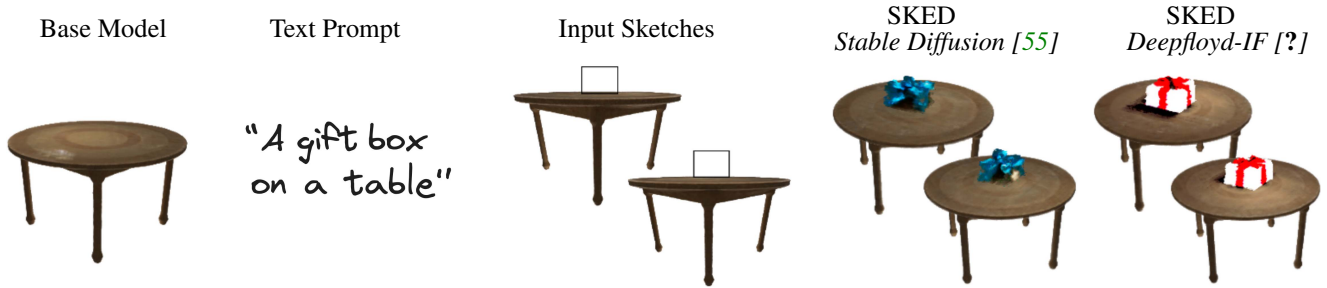
\begin{figure*}[t]

    \centering
    \begin{overpic}[width=1\textwidth,tics=10, trim=0 0 0 0,clip]{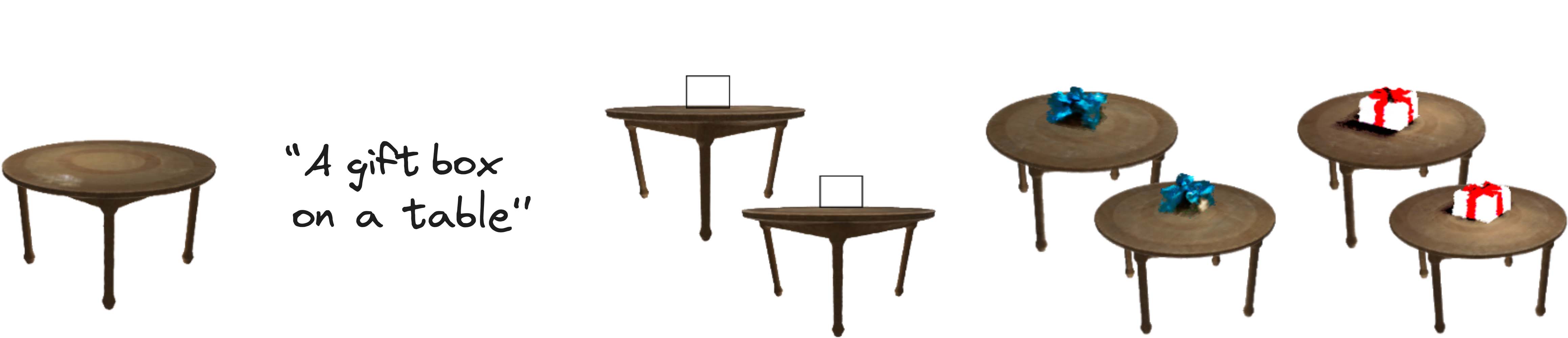}
    \put(2.5,19){Base Model}
    \put(20.0,19){Text Prompt}
    \put(42.0,19){Input Sketches}
    \put(67.0,20){\ours{}}
    \put(63.0,18){\textit{Stable Diffusion \cite{rombach2021highresolution}}}    
    \put(88.25,20){\ours{}}
    \put(85.25,18){\textit{Deepfloyd-IF \cite{deepfloyd}}}
    \end{overpic}
    \captionsetup{font=small}
    \caption{Effect of Diffusion Model Backbone. \ours{} is compatible with any diffusion models applicable with \cite{poole2022dreamfusion}, e.g, the diffusion model should be trained to generate images in the editing domain, and respond to directional prompts, as designated by \cite{poole2022dreamfusion}.}
    \label{fig:diff_diff}
\end{figure*}

To perform sketch-based, local editing, we use two 
loss terms,
$\mathcal{L}_{pres}$, which preserves the base object, and $\mathcal{L}_{sil}$ which generates the desired edit according to the input sketches. We visually analyze the effect of these 
loss terms
on two examples: one reconstructed with \cite{mueller2022instant} (Fig.~\ref{fig:ablations} top row) and another generated by Stable-DreamFusion \cite{stable-dreamfusion} (Fig.~\ref{fig:ablations} bottom row). We differentiate between these two examples as editing neural fields generated by Stable-DreamFusion ensures the rendered base model input is within the diffusion model distribution, 
which leads to fewer adversarial artifacts (see discussion in Section~\ref{subsubsec:basemodeldist}).

Qualitative ablation results are presented in Fig.~\ref{fig:ablations}. Text-Only, is equivalent to applying DreamFusion~\cite{poole2022dreamfusion} initialized with a pretrained NeRF. This method employs neither of the two geometric losses and adheres to the semnatics of the text prompt but drastically alters the base neural field, $\basenerf$. We also experimented with lowering the learning rate to avoid steering from the base model with high gradients, but that did not help in mitigating this effect.

When only $\mathcal{L}_{sil}$ is employed, the sketch regions are edited according to the text-prompts, but the base region also changes drastically. The flower, a generated NeRF, changes more meaningfully compared to the cat. When only $\mathcal{L}_{pres}$ is applied, no explicit constraint exists to respect the sketches. Therefore, our method yields a color artifact in the proximity of the sketch regions. When both constraints are simultaneously applied (our method), the edits respect both sketches and text prompt, and preserve the base NeRF.

We further validate our claim with a quantitative ablation which repeats the experiments in Section~\ref{subsec:quantitative} for the $\mathcal{L}_{sil}$ and $\mathcal{L}_{pres}$ variants (see Table~\ref{tab:psnr} and Table~\ref{tab:ios}). Evidently, both variants are inferior to the full pipeline.



\section{Conclusion, Limitations, and Future Work}
\label{sec:future}
We presented \ours, a NeRF editing method conditioned on text and sketch. Using novel loss functions, our framework allows for local editing of neural fields.
\begin{wrapfigure}{r}{0.2\textwidth} 
\vspace{-10pt}
  \begin{center}
    \includegraphics[width=0.2\textwidth]{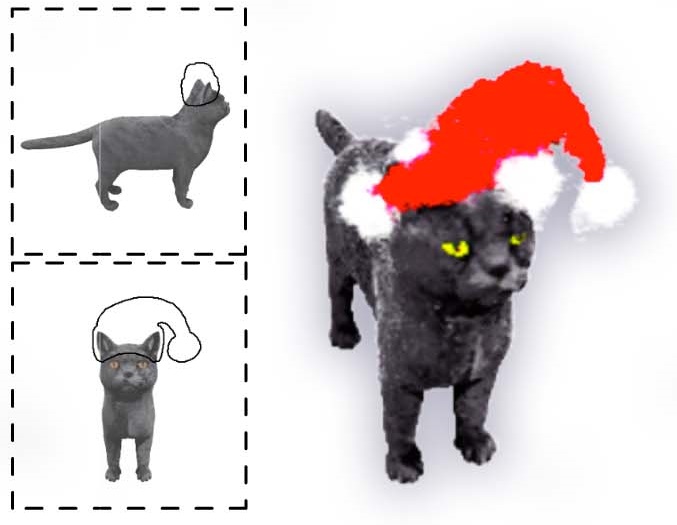}
  \end{center}
    \vspace{-15pt}
 \vspace{1pt}
\end{wrapfigure} 
Similar to previous works \cite{poole2022dreamfusion, lin2022magic3d, metzer2022latent}, our approach utilizes the SDS Loss and may be vulnerable to the well-known "multiface issue" (inset figure) depending on the choice of diffusion model and prompt. Our method supports a single set of prompt and sketch views at a time. A simple workaround is to apply our method multiple times progressively (Fig.~\ref{fig:progressive}). 
Our results rely on the publicly available Stable-Diffusion model \cite{rombach2021highresolution}, which is less amenable to directional text prompts and produces lower quality 3D generated outputs compared to commercial diffusion models used by previous works~\cite{poole2022dreamfusion, lin2022magic3d}. In Fig~\ref{fig:diff_diff} we show that it is possible to get better results by using the Deepfloyd-IF model \cite{deepfloyd}.

Future directions may expand our method to better support for non-opaque materials, or condition on other modalities, possibly through the diffusion model. More research may further extend the usage of sketch scribbles for animation, similar to \cite{dvoro2020monstermash}.




\textbf{Acknowledgements:} We thank Rinon Gal, Masha Shugrina, Roy Bar-On, and Janick Martinez Esturo for proofreading and helpful comments and discussions. This work was funded in part by NSERC Discovery grant (RGPIN-2022-03111), NSERC Discovery Launch Supplement (DGECR-2022-00359), and  the Israel Science Foundation under Grant No. 2492/20.

{\small
\bibliographystyle{ieee_fullname}
\bibliography{egbib}
}

\appendix

\renewcommand\thefigure{\arabic{figure}}
\setcounter{figure}{10}
\renewcommand\thetable{\arabic{table}}
\setcounter{table}{3}


\section{Background}

In the following, we include an extended background chapter cut off from the main paper for brevity.

\subsection{Latent diffusion models (LDMs)}
LDMs 
\cite{rombach2021highresolution}
are a class of diffusion models that operate on a latent space instead of directly sampling high-resolution color images.
These models have two main components: a variational autoencoder consisting of an encoder $\mathcal{E}(x)$ and a decoder $\mathcal{D}(z)$, pretrained on the training data, and a denoising diffusion probabilistic model (DDPM) trained on the latent space of the autoencoder. Specifically, let $Z$ be the latent space learned by the autoencoder.\
The objective of the DDPM is to minimize the following expectation:
\begin{equation}
   \mathbb{E}_{z_0 \sim Z, \epsilon \sim \mathcal{N}(0, I), t}[||\epsilon_\phi(z_t, t) - \epsilon || ^ 2],
\end{equation}
where $t$ is the time-step of the diffusion process, $z_t~=~ \sqrt{\alpha_t}z_0 + \sqrt{1 - \alpha_t}\epsilon$ is the input latent image with noise added to it, and $\epsilon_\phi$ is the denoising model, often constructed as a U-Net~\cite{ho2020ddpm}.
Once trained, it is possible to sample from the latent space $Z$ by starting from a random standard Gaussian noise and running the backward diffusion process as described by Ho et al.~\cite{ho2020ddpm}. 
The sampled latent image then can be fed to $D(z)$ to get a final high-resolution image.

\subsection{Score distillation sampling (SDS)}
\label{subsec:prelim}


First introduced by DreamFusion~\cite{poole2022dreamfusion}, SDS is a method of generating gradients from a pretrained diffusion model, by using its \textit{Score Function} to push the outputs of a parameterized image model towards the mode of the diffusion model distribution. More formally, let $I_\theta$ be an image model with parameters $\theta$. In the case of our application, $I_\theta$ is a neural renderer such as NeRF ~\cite{mildenhall2020nerf} or Instant-NGP ~\cite{mueller2022instant}. We can use a pretrained diffusion model with denoiser $\epsilon_\phi(z_t, t)$, to optimize the following:
\begin{equation}
    \min_{\theta}\mathbb{E}_{\epsilon \sim \mathcal{N}(0, I), t}[||\epsilon_\phi(\sqrt{\alpha_t}I_\theta + \sqrt{1 - \alpha_t}\epsilon, t) - \epsilon|| ^ 2],
\end{equation}
where $t$ is the time-step of the diffusion process, and $\alpha_t$ is a constant scheduling the diffusion forward and backward processes. The Jacobian of the denoiser can be omitted in the gradient of the above expression, to get:
\begin{equation}
    \mathbb{E}_{\epsilon \sim \mathcal{N}(0, I), t}[(\epsilon_\phi(\sqrt{\alpha_t}I_\theta + \sqrt{1 - \alpha_t}\epsilon, t) - \epsilon)\frac{\partial I_\theta}{\partial \theta}].
\end{equation}
The advantage of SDS is that one can apply constraints directly on the image model making this framework suitable for our application of sketch-guided 3D generation.



\section{Additional Evaluations}

\subsection{Quantitative Comparisons}
\begin{table*}
\centering

\captionsetup{font=small}
\caption{
Fidelity of base field. We measure the \textbf{Structural Similarity (SSIM $\uparrow$)} of the method's output against renderings from the base model. \ours  \enspace \textit{(no-preserve)} refers to a variant of our method which doesn't apply $\mathcal{L}_{pres}$. Text-Only refers to a public re-implementation of Latent-NeRF~\cite{metzer2022latent}. Latent-NeRF uses the setting from Section~\ref{sec:supp_qualitative}.
}
\label{tab:ssim}

\begin{tabular}{l \tsm \tsm c \ts c \tsm c \ts c \tsm c \ts c \tsm c \ts c \tsm c \ts c \tsm c}
\toprule

{Method} & 
\multicolumn{2}{c \tsm  \ts}{{Cat}} &
\multicolumn{2}{c \tsm}{{Cupcake}} &
\multicolumn{2}{c \tsm}{{Horse}} &
\multicolumn{2}{c \tsm}{{Sundae}} &
\multicolumn{2}{c \tsm}{{Plant}} &
{{Mean}} \\

& 
\multicolumn{2}{c \tsm  \ts}{{\textit{+chef hat}}} &
\multicolumn{2}{c \tsm}{{\textit{+candle}}} &
\multicolumn{2}{c \tsm}{{\textit{+horn}}} &
\multicolumn{2}{c \tsm}{{\textit{+cherry}}} &
\multicolumn{2}{c \tsm}{{\textit{+flower}}} &
\\

& A & B
& A & B
& A & B
& A & B
& A & B
&
\\
\midrule

\ours  & 
\textbf{0.978} & \textbf{0.990} & 
\textbf{0.964} & \textbf{0.973} & 
\textbf{0.990} & \textbf{0.986} & 
\textbf{0.963} & \textbf{0.962} & 
0.927 & \textbf{0.938} & 
\textbf{0.967}
\\

\ours \enspace \textit{(no-preserve)}  & 
0.867 & 0.890 & 
0.944 & 0.948 & 
0.950 & 0.934 & 
0.913 & 0.921 & 
0.803 & 0.801 & 
0.897
\\

Text-Only~\cite{stable-dreamfusion}  & 
0.875 & 0.918 & 
0.937 & 0.943 & 
0.933 & 0.908 & 
0.947 & 0.951 & 
0.891 & 0.883 & 
0.919
\\

Latent-NeRF~\cite{metzer2022latent} & 
0.915 & 0.948 & 
0.950 & 0.956 & 
0.947 & 0.927 & 
0.904 & 0.906 & 
\textbf{0.930} & 0.925 & 
0.930
\\

\bottomrule

\end{tabular}
\end{table*}

\begin{table*}
\centering

\captionsetup{font=small}
\caption{
Fidelity of base field. We measure the \textbf{Perceptual Image Patch Similarity (LPIPS $\downarrow$)} of the method's output against renderings from the base model. We use VGG \cite{simon2015vgg} as the learned perceptual encoder. \ours  \enspace \textit{(no-preserv)} refers to a variant of our method which doesn't apply $\mathcal{L}_{pres}$. Text-Only refers to a public re-implementation of DreamFusion~\cite{poole2022dreamfusion}. Latent-NeRF uses the setting from Section~\ref{sec:supp_qualitative}.
}
\label{tab:lpips}

\begin{tabular}{l \tsm \tsm c \ts c \tsm c \ts c \tsm c \ts c \tsm c \ts c \tsm c \ts c \tsm c}
\toprule

{Method} & 
\multicolumn{2}{c \tsm  \ts}{{Cat}} &
\multicolumn{2}{c \tsm}{{Cupcake}} &
\multicolumn{2}{c \tsm}{{Horse}} &
\multicolumn{2}{c \tsm}{{Sundae}} &
\multicolumn{2}{c \tsm}{{Plant}} &
{{Mean}} \\

& 
\multicolumn{2}{c \tsm  \ts}{{\textit{+chef hat}}} &
\multicolumn{2}{c \tsm}{{\textit{+candle}}} &
\multicolumn{2}{c \tsm}{{\textit{+horn}}} &
\multicolumn{2}{c \tsm}{{\textit{+cherry}}} &
\multicolumn{2}{c \tsm}{{\textit{+flower}}} &
\\

& A & B
& A & B
& A & B
& A & B
& A & B
&
\\
\midrule

\ours  & 
\textbf{0.070} & \textbf{0.069} & 
0.069 & \textbf{0.061} & 
\textbf{0.028} & \textbf{0.032} & 
0.086 & 0.094 & 
0.158 & 0.128 & 
\textbf{0.079}
\\

\ours \enspace \textit{(no-preserv)}  & 
0.290 & 0.250 & 
0.091 & 0.093 & 
0.089 & 0.098 & 
0.169 & 0.154 & 
0.291 & 0.309 & 
0.183
\\

Text-Only~\cite{stable-dreamfusion}  & 
0.150 & 0.137 & 
0.076 & 0.076 & 
0.115 & 0.134 & 
\textbf{0.081} & \textbf{0.079} & 
0.170 & 0.180 & 
0.120
\\

Latent-NeRF~\cite{metzer2022latent}  & 
0.102 & 0.101 & 
\textbf{0.066} & 0.065 & 
0.081 & 0.100 & 
0.139 & 0.141 & 
\textbf{0.108} & \textbf{0.113} & 
0.101
\\

\bottomrule

\end{tabular}
\end{table*}

 \begin{table*}
\vspace{10mm}
\centering
\captionsetup{font=small}
\caption{
Fidelity of base field. Following the experiments in section~\ref{subsec:quantitative}, we measure the PSNR of the base objects on additional examples provided in Fig.~\ref{fig:additional} and Fig.~\ref{fig:diff_diff}.
}
\label{tab:psnr_supp}

\begin{tabular}{@{}llrlrlrlrlr@{}}
\toprule
& \multicolumn{2}{c}{Tree to Cactus} & \multicolumn{2}{c}{Anime+Skirt} & \multicolumn{2}{c}{Pancake+Cream }  & \multicolumn{2}{c}{Gift on Table} & \multicolumn{2}{c}{Mean}  \\ 
\cmidrule(l){2-3} \cmidrule(l){4-5} \cmidrule(l){6-7} \cmidrule(l){8-9} \cmidrule(l){10-10}
Method & view 1 & view 2 & view 1 & view 2 & view 1 & view 2 & view 1 & view 2 &    \\ \midrule
SKED & \textbf{29.15} & \textbf{27.47} & \textbf{39.67} & \textbf{37.40} & \textbf{27.48} & \textbf{26.64} & \textbf{34.16} & \textbf{31.52}  & \textbf{31.68}\\
 Text-Only & 23.12 & 24.40 & 22.61 & 21.95 & 16.97 & 15.35 & 19.05 & 20.70 & 20.51 \\

\bottomrule
\end{tabular}

\label{tab:reb:ablation}
\vspace{20mm}
\end{table*}

\textbf{Base Model Fidelity.} In Table~\ref{tab:ssim}, We include the SSIM metric to further quantify our method's capability to preserve the base model.

\subsection{Qualitative Comparisons}
\label{sec:supp_qualitative}


\begin{figure*}
    \centering
    \includegraphics[width=0.8\linewidth]{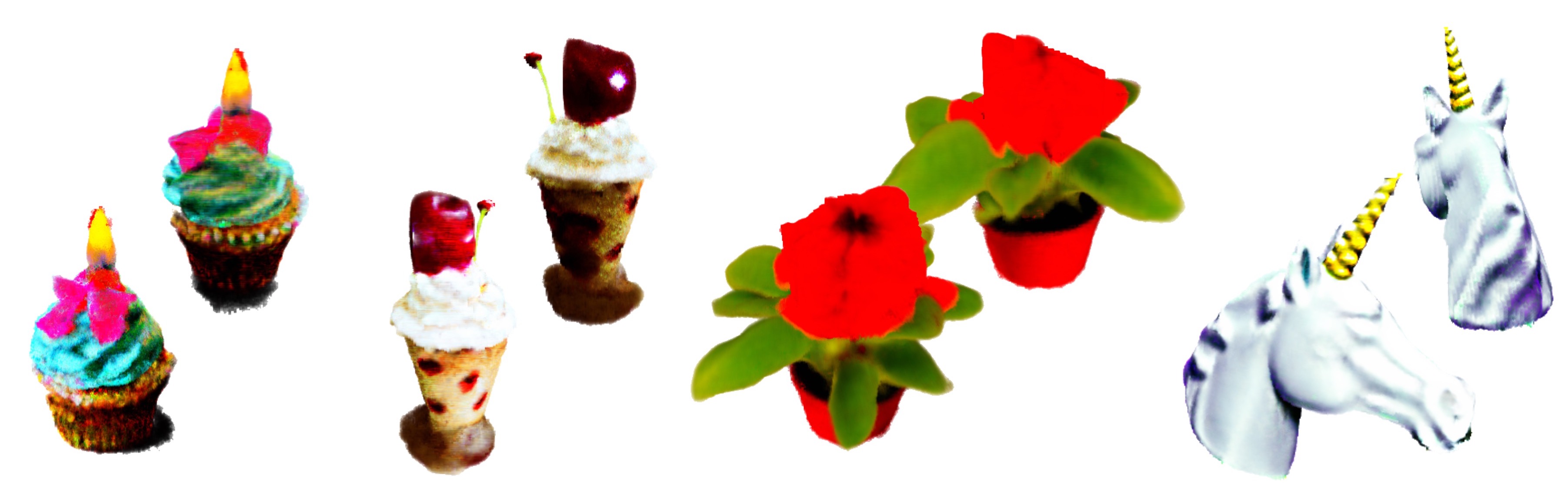}
    \captionsetup{font=small}
    \caption{Examples from the modified version of the sketch shape pipeline of Latent-NeRF~\cite{metzer2022latent}}
    \label{fig:comparison}
\end{figure*}











\begin{table*}
\centering
\setlength{\tabcolsep}{4pt}

\captionsetup{font=small}
\caption{
To compare our method's ability to preserve the base with the baseline derived from Latent-NeRF~\cite{metzer2022latent}, we measure the PSNR of both method's outputs against renderings from the base model. Additionally, we report the average runtime of our method compared to the baseline.
}
\label{tab:compare}

\begin{tabular}{l \tsm \tsm c \ts c \tsm c \ts c \tsm c \ts c \tsm c \ts c \tsm c \ts c \tsm c \tsm c}
\toprule

{Method} & 
\multicolumn{2}{c \tsm  \ts}{{Cat}} &
\multicolumn{2}{c \tsm}{{Cupcake}} &
\multicolumn{2}{c \tsm}{{Horse}} &
\multicolumn{2}{c \tsm}{{Sundae}} &
\multicolumn{2}{c \tsm}{{Plant}} &
{{PSNR}} &
{{Runtime (minutes)}}\\

& 
\multicolumn{2}{c \tsm  \ts}{{\textit{+chef hat}}} &
\multicolumn{2}{c \tsm}{{\textit{+candle}}} &
\multicolumn{2}{c \tsm}{{\textit{+horn}}} &
\multicolumn{2}{c \tsm}{{\textit{+cherry}}} &
\multicolumn{2}{c \tsm}{{\textit{+flower}}} &
Mean &
Mean
\\

& A & B
& A & B
& A & B
& A & B
& A & B
&
\\
\midrule

\ours  & 
\textbf{31.05} & \textbf{34.13} & 
\textbf{23.73} & \textbf{25.98} & 
\textbf{32.45} & \textbf{31.46} & 
\textbf{26.47} & \textbf{25.99} & 
\textbf{21.71} & \textbf{22.31} & 
\textbf{27.53} &
\textbf{38}
\\

Latent-NeRF~\cite{metzer2022latent}  & 
21.15 & 22.62& 
21.99 & 21.20 & 
17.00 & 15.97 & 
16.07 & 15.47 & 
17.66 & 16.78 & 
18.59 &
64
\\

\bottomrule

\end{tabular}
\end{table*}

\textbf{Comparison to Latent-NeRF~\cite{metzer2022latent}.} To the best of our knowledge, we are the first work to employ 2D sketch-based editing of NeRFs. Given that prior works are not directly comparable with our editing setting, we attempt to create a close comparison instead, faithful to the original compared method and fair to evaluate our editing setting. As baseline, we use the method from Latent-NeRF's
~\cite{metzer2022latent} 
3D sketch shape pipeline. We initialize a NeRF with the base object weights, and create a \textit{3D sketch shape}, a mesh, by intersecting the bounding boxes of our 2D sketches in the 3D space. Note that we could also intersect the sketch masks, however, due to view inconsistencies, we found that the results are far inferior. After initializing the NeRF and creating the sketch shape, we proceed to use the sketch shape loss from the paper to preserve the geometry, while editing the NeRF according to the input text. In Fig.~\ref{fig:comparison}, we establish that while this baseline is able to perform meaningful edits, it suffers from two apparent issues: (i) the baseline severely changes the base NeRF, and (ii) the edited region is bound to the coarse geometry of the intersected bounding boxes. To alleviate the latter, one could resort to modeling 3D assets as a sketch shape. However, we show that by using simple multiview sketches, it is possible to perform local editing without going through the effort of modeling accurate 3D masks. Finally, we include a quantitative summary of the preservation ability and the performance of the two methods in Table~\ref{tab:compare}.




\begin{figure*}[t]
   \centering
    \begin{overpic}[width=0.8\linewidth,tics=10, trim=0 0 0 0,clip]{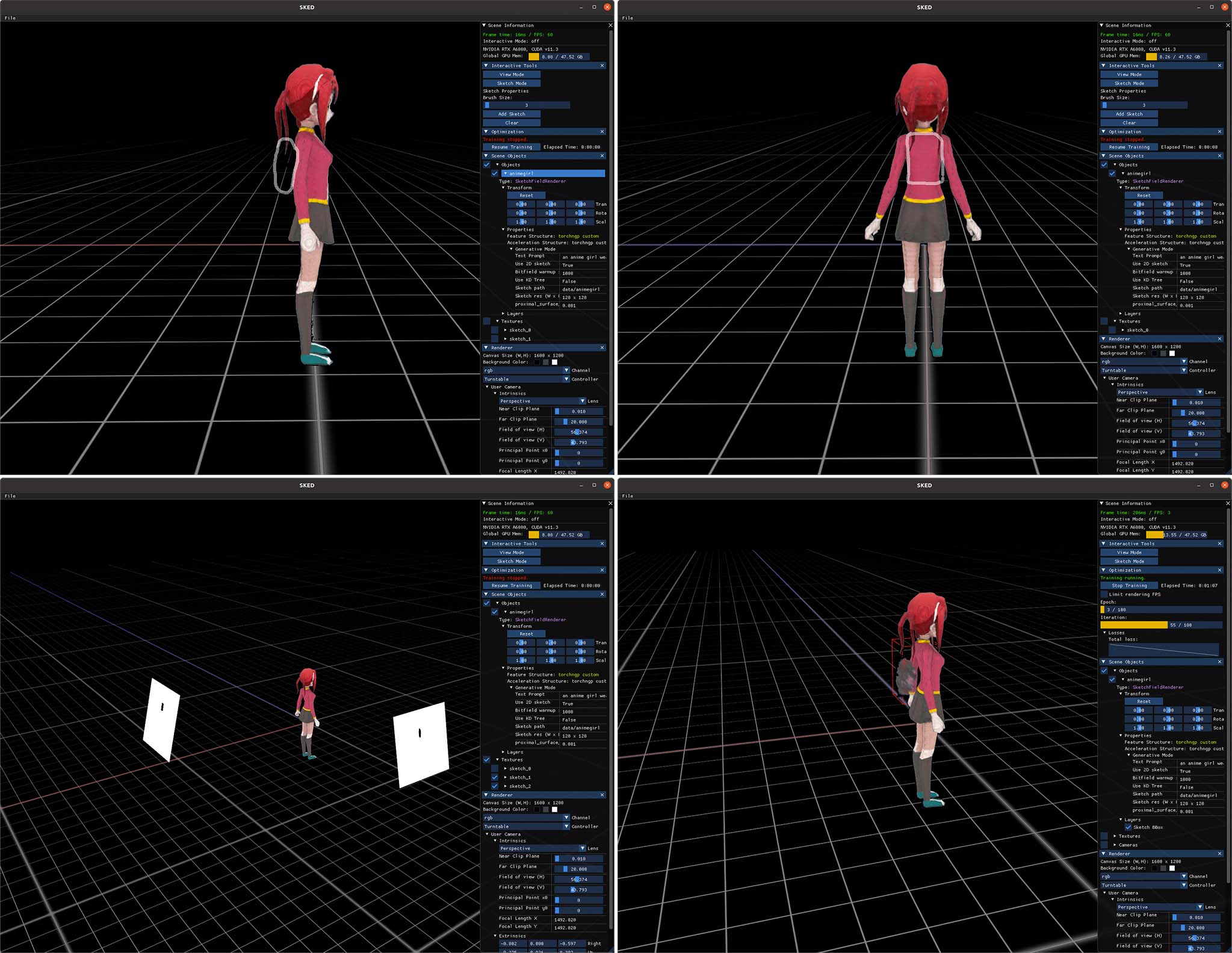}
   \end{overpic}
   \captionsetup{font=small}
    \caption{The interactive UI allows users to sketch over a pretrained NeRF. \textbf{Top row}: The user draws scribbles from two different views using "Sketch Mode". \textbf{Bottom left}: After pressing "Add sketch", the scribbles are filled to generate masks, ready to be used with our pipeline. \textbf{Bottom right}: The bounding box marks the sketches intersection region, where the edit takes place.}
    \label{fig:ui}
 \end{figure*}

 \begin{figure*}[h]
    \centering
    \hspace{0.25in}
    \begin{overpic}[width=0.8\linewidth,tics=10, trim=0 0 0 0,clip]{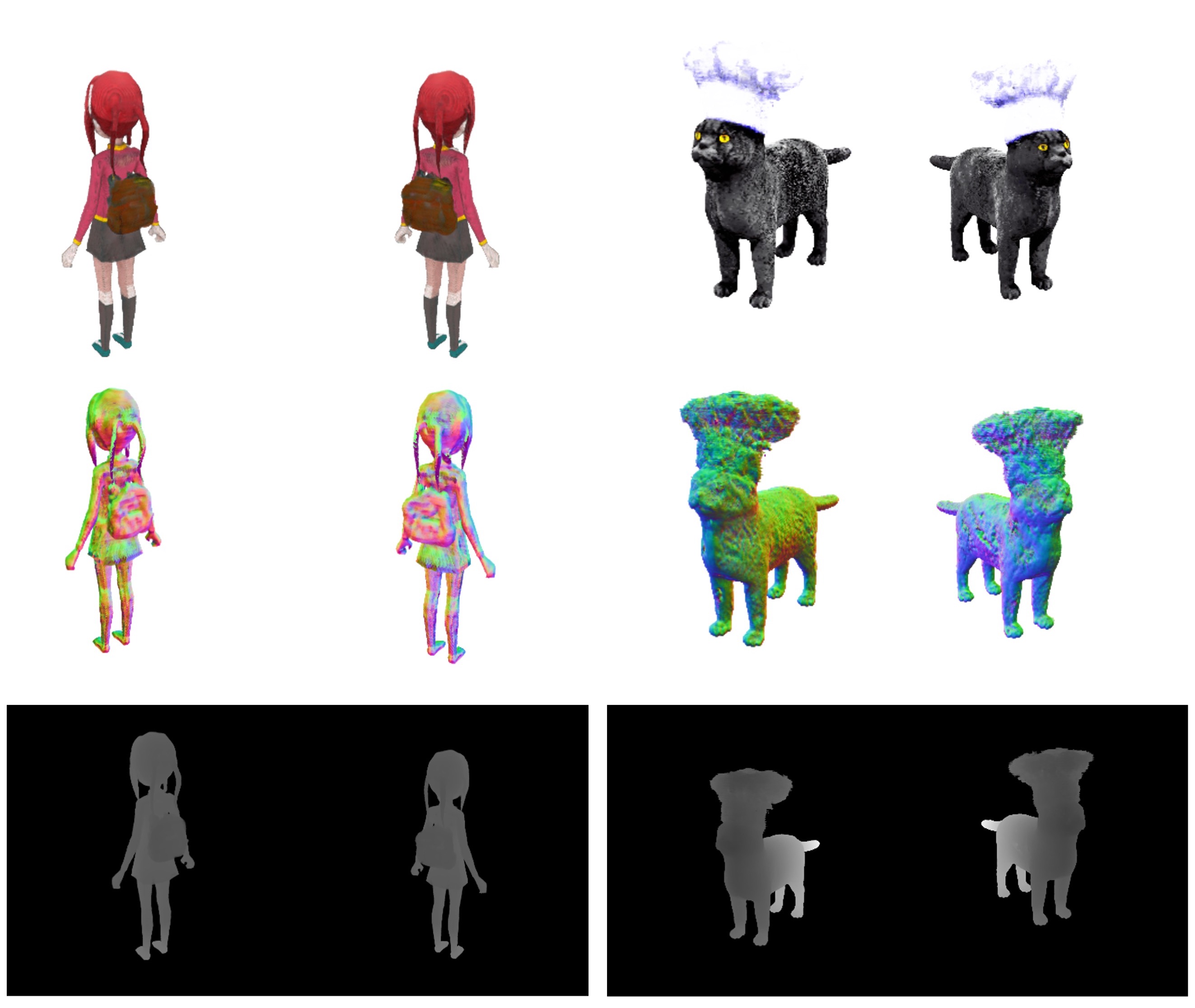}
    \end{overpic}
    \captionsetup{font=small}
    \caption{From top to bottom: color, normal and depth maps of outputs generated by our method.}
    \label{fig:geometry}
\end{figure*}

\section{Implementation Details}
\vspace{-10pt}
This section contains additional implementation details omitted from the manuscript.
\vspace{-35pt}
\subsection{Text Prompts}
\vspace{-10pt}
In the following, we include the full list of prompts that were used to generate the examples within the paper.
\vspace{-40pt}
\begin{itemize}
\setlength{\parskip}{0pt}
  \setlength{\itemsep}{0pt}
    \item "A cat wearing a chef hat"
    \item "A cherry on top of a sundae"
    \item "A red flower stem rising from a potted plant"
    \item "A teddy bear wearing sunglasses"
    \item "A candle on top of a cupcake"
    \item "An anime girl wearing a brown bag"
    \item "An apple on a plate"
    \item "A Nutella jar on a plate"
    \item "A globe on a plate"
    \item "A tennis ball on a plate"
    \item "A cat wearing a red tie"
    \item "A cat wearing red tie wearing a chef hat"
    \item "A 3D model of a unicorn head"
\end{itemize}
\vspace{-40pt}

Additionally, similar to 
DreamFusion\cite{poole2022dreamfusion} 
we use directional prompts, where based on the rendering view, we modify prompt $\textbf{T}$ as follows:
\vspace{-40pt}
\begin{itemize}
\setlength{\parskip}{0pt}
  \setlength{\itemsep}{0pt}
    \item "$\textbf{T}$, overhead view"
    \item "$\textbf{T}$, side view"
    \item "$\textbf{T}$, back view"
    \item "$\textbf{T}$, bottom view"
    \item "$\textbf{T}$, front view"
\end{itemize}
\vspace{-60pt}



\subsection{Interactive UI}
\vspace{-\baselineskip}

Since our method requires user interaction, we include an interactive user interface with our implementation (Fig.~\ref{fig:ui}). The user interface allows users to optimize newly reconstructed base NeRF models, or load pretrained ones. To perform edits, users can position the camera on the desired sketch view, and draw scribbles to guide SKED. By pressing "Add Sketch", scribbles are filled and converted to masked sketch inputs, ready to be used with our method.
\vspace{-30pt}
\subsection{Quality Notes}

Our implementation uses an early version of Stable-DreamFusion
~\cite{stable-dreamfusion}
which does not include the optimizations very recently suggested by
Magic3D~\cite{lin2022magic3d}. 
In contrast to DreamFusion
~\cite{poole2022dreamfusion} 
and Magic3D
~\cite{lin2022magic3d},
which use commercial diffusion models with larger language models 
\cite{imagen2022saharia, balaji2022eDiff-I},
we rely on Stable Diffusion
\cite{rombach2021highresolution}, 
which is less sensitive to directional prompts. Our results are therefore not comparable in visual quality to these previous works.


\section{Additional Assets}

\subsection{Geometry and Depth}
In addition to RGB images, we share examples highlighting the geometry of our method's outputs. In Fig.~\ref{fig:geometry} we include the normal maps and depth maps of two output samples.



\end{document}